\documentclass[11pt,a4paper]{article}
\usepackage[hyperref]{acl2017}
\usepackage{times}
\usepackage{latexsym}
\usepackage{graphicx}

\usepackage{url}
\usepackage{amssymb}
\usepackage{amsmath}
\usepackage{amsfonts}

\aclfinalcopy 

\title{A Survey of Deep Learning Methods for Relation Extraction}

\author{Shantanu Kumar\\
  Indian Institute of Technology Delhi \\
  {\tt ee1130798@iitd.ac.in} \\}

\date{}

\begin{document}
\maketitle

\begin{abstract}
Relation Extraction is an important subtask of Information Extraction which has the potential of employing deep learning (DL) models with the creation of large datasets using distant supervision. In this review, we compare the contributions and pitfalls of the various DL models that have been used for the task, to help guide the path ahead.
\end{abstract}

\section{Introduction}

Information Extraction (IE) is a task of natural language processing that involves extracting structured information, that can be interpreted easily by a machine or a program, from plain unstructured text. Since the Internet is filled with huge amounts of data in the form of text, IE systems are extremely important. They can extract meaningful facts from this text, which can then be used for applications like search and QA. Knowledge-bases like Freebase \citep{bollacker2008freebase} and DBpedia \citep{auer2007dbpedia} which are a source for useful information are far from complete and can be extended using such systems. Information Extraction itself is a huge task consisting of several subtasks like named-entity-recognition, relation extraction, event extraction etc. In this review, we specifically focus on deep learning methods used for the subtask of relation extraction. 

IE can be done in unsupervised or semi-supervised domain, in the form of OpenIE, where we do not have any predefined ontology or relation classes and we extract facts from the data along with the relation phrases. In the supervised domain, the relation extraction and classification tasks specifically refers to the classification of an entity pair to a set of known relations, using documents containing mentions of the entity pair. The RE task refers to predicting whether a given document contains a relation or not for the pair, modeled as a binary classification. Relation classification refers to predicting which relation class out of a given ontology does that document point to, given that it does contain a relation (modeled as a multi-class classification problem). The two tasks can be combined by making a multi-class classification problem with an extra \textit{NoRelation} class.

Traditional non deep learning methods for relation extraction typically work in the supervised paradigm. They can be divided into two classes which are feature based methods and kernel based methods. In both these methods, the extracted features and elaborately-designed kernels use pre-existing NLP systems which result in errors of the various modules accumulating downstream. Also, the manually constructed features may not capture all the relevant information that is required. This need to manual engineer features is removed by  moving into the domain of deep learning. 

Supervised techniques for machine learning require large amount of training data for learning. Using hand annotated datasets for relation extraction takes huge time and effort to make the datasets. \citet{mintz2009distant} proposed a distant supervision method for producing large amount of training data by aligning KB facts with texts. Such large datasets allow for learning more complex models for the task like convolutional neural networks. The noise present in datasets generated through distant supervision also require special ways of modeling the problem like Multi-Instance Learning as discussed in the subsequent sections. 


\section{Datasets}

\subsection{Supervised Training}

The early works on relation extraction using deep learning employed supervised training datasets that were previously used by non deep learning models. These datasets required intensive human annotation which meant that the data contained high quality tuples with little to no noise. But human annotation can be time-consuming, as a result of which these datasets were generally small. Both of the datasets mentioned below contain data samples in which the document sentence is already labeled with named entities of interest and the relation class expressed between the entity pair is to be predicted.

\begin{description}
\item[ACE 2005 dataset] The Automatic Content Extraction dataset contains 599 documents related to news and email and contains relations that are divided into 7 major types. Out of these, 6 major relation types contain enough instances (average of 700 instances per relation type) and are used for training and testing.
\item[SemEval-2010 Task 8 dataset] This dataset is a freely available dataset by \citet{hendrickx2009semeval} which contains 10,717 samples which are divided as 8,000 for training and 2,717 for testing. It contains 9 relation types which are ordered relations. The directionality of the relations effectively doubles the number of relations, since an entity pair is believed to be correctly labeled only if the order is also correct. The final dataset thus has 19 relation classes (2 $\times$ 9 + 1 for \textit{Other} class). 
\end{description}

\subsection{Distant Supervision}

To avoid the laborious task of manually building datasets for relation extraction, \citet{mintz2009distant} proposed a distant supervision approach for automatically generating large amounts of training data. They aligned documents with known KBs, using the assumption that if a relation exists between an entity pair in the KB, then every document containing the mention of the entity pair would express that relation. It can easily be realised that this distant supervision assumption is a very strong assumption and that every document containing the entity pair mention may not express the relation between the pair. Eg. For the tuple (\textit{Microsoft}, \texttt{Founder\_of}, \textit{Microsoft}) in the database and the document ``\textit{Bill Gates’s turn to philanthropy was linked to the antitrust problems Microsoft had in the U.S. and the European union}", the document does not express the relation \texttt{Founder\_of} even though it contains both the entities.

To alleviate this problem and reduce the noise, \citet{riedel2010modeling} relaxed the distant supervision assumption by modeling the problem as a multi-instance learning problem (described in the subsequent section). The dataset they used is the most popular dataset used in subsequent works building on distant supervision for relation extraction. This dataset was formed by aligning Freebase relations with the New York Times corpus (NYT). Entity mentions were found in the documents using the Stanford named entity tagger, and are further matched to the names of Freebase entities. There are 53 possible relation classes including a special relation \textit{NA} which indicates there is no relation between the entity pair. The training data contains 522,611 sentences, 281,270 entity pairs and 18,252 relational facts. The testing set contains 172,448 sentences, 96,678 entity pairs and 1,950 relational facts.

The evaluation for this dataset is usually done by comparing the extracted facts against the entries in Freebase. However, since Freebase is not a complete KB, the evaluation scheme is affected by false negatives that undermine the performance of the models. For a comparative study, however, the evaluation scheme works alright.

\section{Basic Concepts}
The following section talks about some basic concepts that are common across most deep learning models for relation extraction.

\subsection{Word Embeddings}
Word embeddings \citep{mikolov2013distributed,pennington2014glove} are a form of distributional representations for the words in a vocabulary, where each word is expressed as a vector in a low dimensional space (low w.r.t to the size of the vocabulary). Word embeddings aim to capture the syntactic and semantic information about the word. They are learnt using unsupervised methods over large unlabeled text corpora. They are implemented using an embedding matrix $E \in \mathbb{R}^{|V| \times d_w}$, where $d_w$ is the dimensionality of the embedding space and $|V|$ is the size of the vocabulary.

\subsection{Positional Embeddings}
In the relation extraction task, along with word embeddings, the input to the model also usually encodes the relative distance of each word from the entities in the sentence, with the help of positional embeddings (as introduced by \citet{zeng2014relation}). This helps the network to keep track of how close each word is to each entity. The idea is that words closer to the target entities usually contain more useful information regarding the relation class. The positional embeddings comprise of the relative distance of the current word from the entities. For example, in the sentence ``Bill\_Gates is the founder of Microsoft.", the relative distance of the word ``founder" to head entity ``Bill\_Gates" is 3 and tail entity ``Microsoft" is -2. The distance are then encoded in a $d_p$ dimensional embedding.

Finally, the overall sentence $x$ can expressed as a sequence of vectors $x = \{w_1, w_2, ..., w_m\}$ where every word $w_i \in \mathbb{R}^{d}$ and $d = d_w + 2\times d_p$. 

\subsection{Convolutional Neural Networks}

For encoding the sentences further, deep learning models for relation extraction usually use convolutional neural network layers to capture n-gram level features, similar to \citet{collobert2011natural}. The convolutional layer operates as follows. Given an input sentences $x$ as a sequence of vectors $x = \{w_1, w_2, ..., w_m\}, w_i \in \mathbb{R}^d$, if $l$ is the window size for the convolutional layer kernel, then the vector for the $i$-th window ($q_i \in \mathbb{R}^{(d\times l)}$) is formed by concatenating the input vectors for that window,

\begin{equation}
q_i = w_{i:i+l-1} ; (1\leq i\leq m-l+1)
\end{equation}

A single convolutional kernel would then consist of a weight vector $W \in \mathbb{R}^{(d\times l)}$ and a bias $b \in \mathbb{R}$, and the output for the $i$-th window is computed as,

\begin{equation}
p_i = f(W'q_i + b)
\end{equation}

where $f$ is the activation function. Hence the output of the convolutional kernel $p$ would be of the shape $p \in \mathbb{R}^{(m-l+1)}$. A convolutional layer can consist of $d_c$ convolutional kernels which would make the output of the convolutional layer of the shape $\mathbb{R}^{d_c \times (m-l+1)}$.

\begin{figure}[ht]
	\centering
	\includegraphics[width=0.5\textwidth]{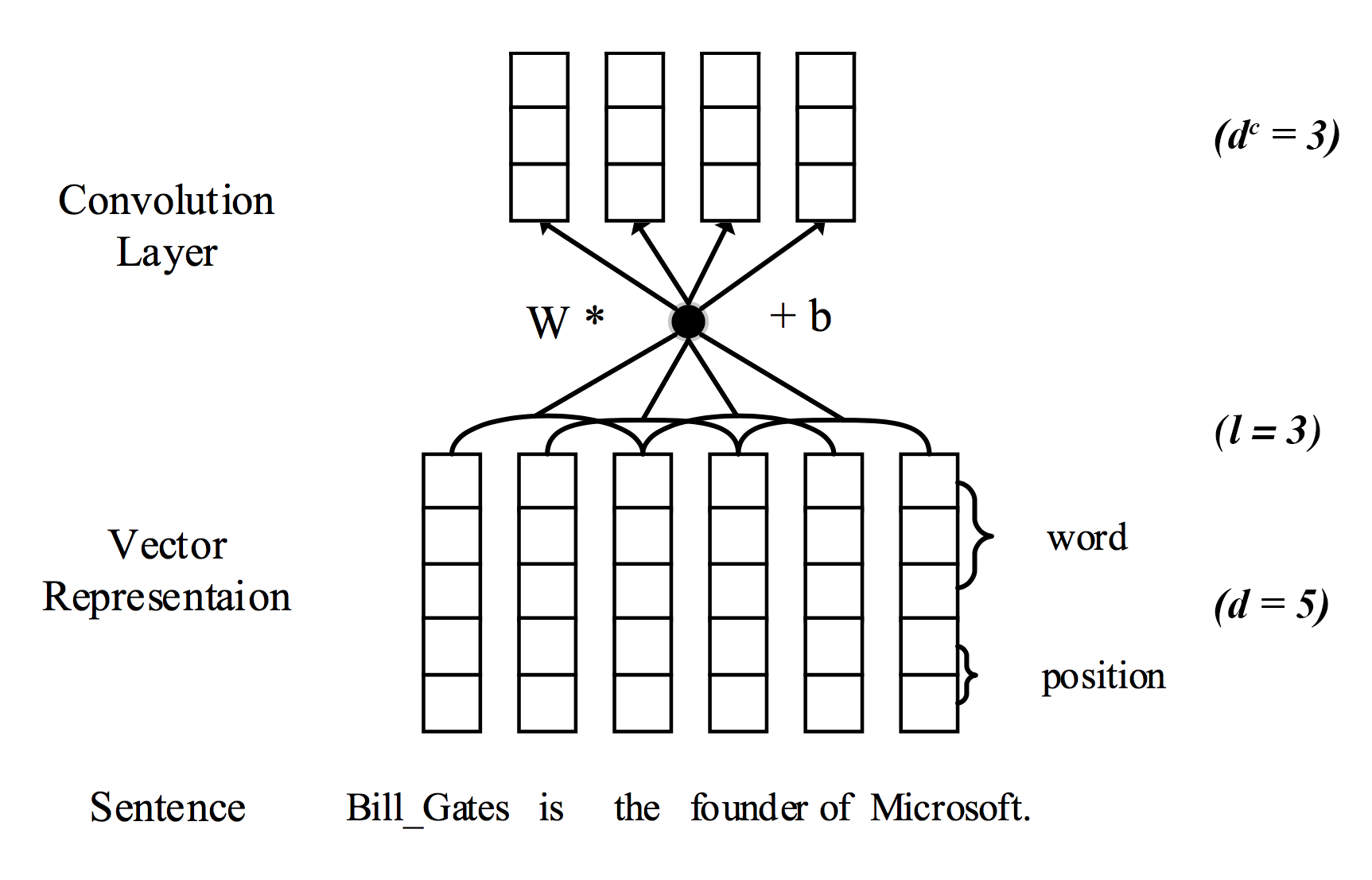}
    \caption{Encoder structure with Word and Positional Embeddings followed by Convolutional Layer. (Sourced from \citep{lin2016neural})}
\end{figure}


\begin{table*}
\centering
\begin{tabular}{cccccc}

\textbf{Model} & \textbf{\begin{tabular}[c]{@{}c@{}}Multi-instance \\ Learning\end{tabular}} & \textbf{\begin{tabular}[c]{@{}c@{}}Word \\ Embeddings\end{tabular}} & \textbf{\begin{tabular}[c]{@{}c@{}}Positional \\ Embeddings\end{tabular}} & \textbf{\begin{tabular}[c]{@{}c@{}}Additional \\ Lexical \\ Features\end{tabular}} & \textbf{Max Pooling}                                                    \\ \hline \hline
\citet{liu2013convolution}              & No                                                                          & Random                                                              & No                                                                        & Yes                                                                             & No                                                                      \\ \hline
\citet{zeng2014relation}              & No                                                                          & Pretrained                                                          & \begin{tabular}[c]{@{}c@{}}Yes\\ (Not Trained)\end{tabular}               & Yes                                                                             & Yes                                                                     \\ \hline
\begin{tabular}[c]{@{}l@{}}Nguyen and\\ Grishman (2015)\end{tabular}              & No                                                                          & Word2Vec                                                            & Yes                                                                       & No                                                                              & Yes                                                                     \\ \hline
\begin{tabular}[c]{@{}c@{}}PCNN \\ \citet{zeng2015distant}\end{tabular}              & \begin{tabular}[c]{@{}c@{}}Yes \\ (1 sentence \\ per bag)\end{tabular}                                                                         & Word2Vec                                                            & Yes                                                                       & No                                                                              & \begin{tabular}[c]{@{}c@{}}Yes\\ (Piecewise in \\ a sentence)\end{tabular} \\ \hline
\begin{tabular}[c]{@{}c@{}}PCNN + Att \\ \citet{lin2016neural}\end{tabular}              & \begin{tabular}[c]{@{}c@{}}Yes \\ (Attention weighted \\ sum over bag)\end{tabular}                                                                         & Word2Vec                                                            & Yes                                                                       & No                                                                              & \begin{tabular}[c]{@{}c@{}}Yes\\ (Piecewise \\ and Full)\end{tabular}      \\ \hline
\begin{tabular}[c]{@{}c@{}}MIMLCNN \\ \citet{jiangrelation}\end{tabular}              & \begin{tabular}[c]{@{}c@{}}Yes \\ (Max of each \\ feature over bag)\end{tabular}                                                                        & Word2Vec                                                            & Yes                                                                       & No                                                                              & \begin{tabular}[c]{@{}c@{}}Yes\\ (Cross sentence \\ in a bag)\end{tabular}
\end{tabular}
\caption{Summary of features used in the various models for relation extraction using CNNs}
\end{table*}

\section{Supervised learning with CNNs}
The early works using deep learning for relation extraction worked in the supervised training paradigm with the hand-annotated corpus mentioned previously. These model tried to assigned a relation class label to each sentence containing a mention of the entity pair in focus, by modeling the problem as a multi-class classification problem.

\subsection{Simple CNN model ~\citep{liu2013convolution}}
This work is perhaps the earliest work that tries to use a CNN to automatically learn features instead of hand-craft features. It builds an end-to-end network which first encodes the input sentence using word vectors and lexical features, which is followed by a convolutional kernel layer, a single layer neural network and a softmax output layer to give a probability distribution over all the relation classes.

The model uses synonym vectors instead of word vectors, by assigning a single vector to each synonym class rather than giving every individual word a vector. However, it fails to exploit the real representational power of word embeddings. The embeddings are not  trained in an unsupervised fashion on the corpus, but randomly assigned to each synonym class. Further, the model also tries to incorporate some lexical features using word lists, POS lists and entity type lists. It is found that this model outperforms the state-of-the-art kernel-based model at the time on the ACE 2005 dataset by 9 points of F-score. There were several improvements that could be made in this model, but as primary step it worked as a proof of concept that deep learning models could perform as good or even better than the rigorously engineered feature-based or kernel-based models.

\subsection{CNN model with max-pooling ~\citep{zeng2014relation}}
Similar to the previous model, this paper used a CNN for encoding the sentence level features. But unlike the previous paper, they used word embeddings that were pre-trained on a large unlabeled corpus. The paper was also the first work that used Positional Embeddings described in the earlier section, which were adapted as standard in all subsequent RE deep learning models. This model also used lexical level features like information about the nouns in the sentence and the WordNet hypernyms of the nouns. 

One important contribution of this model was the use of a max-pooling layer over the output of the convolutional network. The output of the convolutional layer $Z \in \mathbb{R}^{d_c\times (m-l+1)}$ is dependent on the size of the input sentence $m$. To make this output independent of $m$ and to capture most useful feature in each dimension of the feature vector across the entire sentence, it was motivated to use a max operation that would collapse $Z$ to $Z' \in \mathbb{R}^{d_c}$. Hence, the the dimension of $Z'$ is no longer related to the sentence length $m$. The model was shown to outperform SVM and MaxEnt based models that used a variety of lexical features. Their ablation study also showed that the Positonal Embeddings gave almost a 9\% improvement in their F-score.

\subsection{CNN with multi-sized window kernels ~\citep{nguyen2015relation}}

This work was one of the last works in supervised domain for relation extraction which built upon the works of \citet{liu2013convolution} and \citet{zeng2014relation}. The model completely gets rid of exterior lexical features to enrich the representation of the input sentence and lets the CNN learn the required features itself. Their architecture is similar to \citet{zeng2014relation} consisting of word and positional embeddings followed by convolution and max-pooling. Additionally, they also incorporate convolutional kernels of varying window sizes to capture wider ranges of n-gram features. By experimenting with different iteration, they find that using kernels with 2-3-4-5 window lengths, gives them the best performance. The authors also initialize the word embedding matrix using pre-trained word embeddings trained with word2vec \citep{mikolov2013distributed}, which gives them a significant boost over random vectors and static-word2vec vectors.

\section{Multi-instance learning models with distant supervision}
As mentioned previously, \citet{riedel2010modeling} relaxed the distant supervision assumption by modeling the task as a multi-instance learning problem, so that they could exploit the large training data created by distant supervision while being robust to the noise in the labels. Multi-instance learning is a form of supervised learning where a label is given to a bag of instances, rather than a single instance. In the context of RE, every entity pair defines a bag and the bag consists of all the sentences that contain the mention of the entity pair. Instead of giving a relation class label to every sentence, a label is instead given to each bag of the relation entity. \citet{riedel2010modeling} model this using the assumption that "if a relation exists between an entity pair, then at least one document in the bag for the entity pair must reflect that relation". 

\subsection{Piecewise Convolutional Neural Networks ~\citep{zeng2015distant}}

The PCNNs model uses the multi-instance learning paradigm, with a neural network model to build a relation extractor using distant supervision data. The neural network architecture is similar to the models by \citep{zeng2014relation} and \citep{nguyen2015relation} discussed previously, with one important contribution of piecewise max-pooling across the sentence. The authors claim that the max-pooling layer drastically reduces the size of the hidden layer and is also not sufficient to capture the structure between the entities in the sentence. This can be avoided by max-pooling in different segments of the sentence instead of the entire sentence. It is claimed that every sentence can naturally be divided into 3 segments based on the positions of the 2 entities in focus. By doing a piecewise max-pool within each of the segments, we get a richer representation while still maintaining a vector that is independent of the input sentences length.

One of the drawbacks in this model which is later addressed in future models is the way in which the multi-instance problem was set in the loss function. The paper defined the loss for training of the model as follows. Given $T$ bags of documents with each bag containing $q_i$ documents and having the label $y_i$, $i = 1,2..,T$, the neural network gives the probability of extracting relation $r$ from document $j$ of bag $i$, $d_i^j$ denoted as,

\begin{equation}
p(r|d_i^j, \theta); j = 1,2,...,q_i
\end{equation}

where $\theta$ is the weight parameters of the neural network. Then the loss is given as,

\begin{equation}
J(\theta) = \sum_{i=1}^T \text{log} p(y_i|d_i^{j^*}, \theta)
\end{equation}

\begin{equation}
j^* = \text{arg} \text{max}_j p(y_i|d_i^j, \theta); j=1,2...,q_i
\end{equation}

Thus, since the method assumes that ``atleast one document in the bag expresses the relation of the entity pair'' it uses only the one most-likely document for the entity pair during the training and prediction stage. This means that the model is neglecting large amounts of useful data and information that is expressed by the other sentences in the bag. Even though not all the sentences in the bag express the correct relation between the entity pair, using only a single sentence is a very hard constraint. This issue is addressed in the subsequent works.

The PCNNs model with Multi-instance learning is shown to outperform the traditional non deep learning models like the distant-supervision based model by \citet{mintz2009distant}, the multi instance learning method \textit{MultiR} proposed by \citet{hoffmann2011knowledge} and the multi-instance multi-label model \textit{MIML} by \citet{surdeanu2012multi}, on the dataset by \citet{riedel2010modeling} (Figure 3). The results are further discussed in the later section. Their ablation study also shows the advantages of using PCNNs over CNNs and Multi-instance learning over traditional learning, which both add incrementally to the model as shown in Figure 2.

\begin{figure}[ht]
	\centering
	\includegraphics[width=0.5\textwidth]{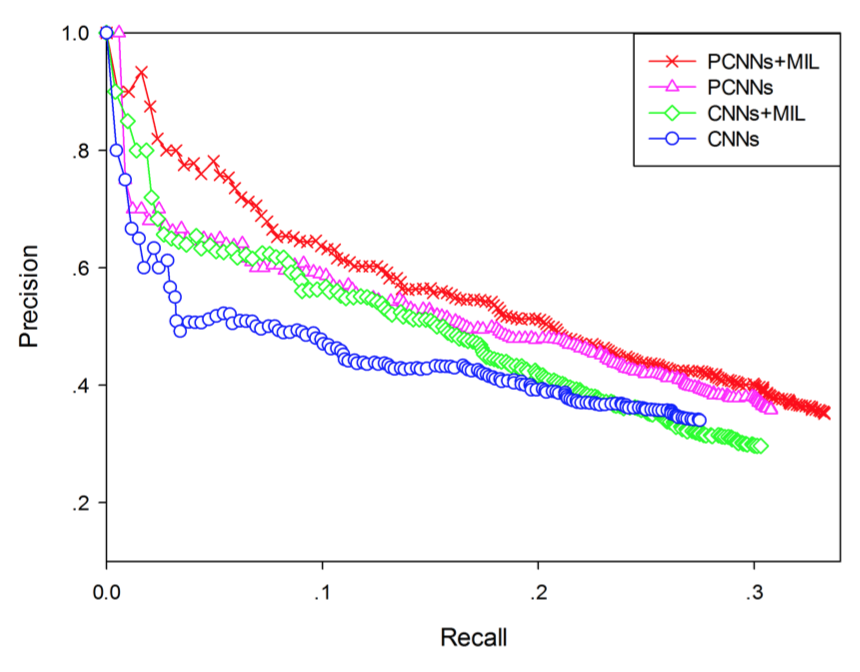}
    \caption{Effect of piecewise max pooling and multi-instance learning. (Sourced from \citep{zeng2015distant})}
\end{figure}

\subsection{Selective Attention over Instances ~\citep{lin2016neural}}

To address the shortcoming of the previous model which only used the one most-relevant sentence from the bag, \citet{lin2016neural} used an attention mechanism over all the instances in the bag for the multi-instance problem. In this model, each sentence $d_i^j$ of bag $i$ is first encoded into a vector representation, $r_i^j$ using a PCNN or a CNN, as defined previously. Then the final vector representation for the bag of sentences is found by taking an attention-weighted average of all the sentence vectors ($r_i^j, j=1,2...q_i$) in the bag. The model computes a weight $\alpha_j$ for each instance $d_i^j$ of bag $i$. These $\alpha$ values are dynamic in the sense that they are different for each bag and depend on the relation type and the document instance. The final feature vector for the bag of sentences is given as,

\begin{equation}
r_i = \sum_{j=1}^{q_i}\alpha_j r_i^j
\end{equation}

When the loss is found using this attention weighted representation of all the instances in the bag, the model is able to inherently identify the important sentences from the noisy ones and all the information in the bag is utilized to make the relation class prediction. 

It can also be observed that the `only-one most likely sentence' model used in the PCNN paper is a special case of the selective attention procedure where $\alpha_{j^*} = 1$ for only $j^*$ as defined by equation (5) and all the remaining $\alpha$ values are zero (hard attention). It is shown that using this selective attention procedure significantly improves the precision recall curve of both the CNN and the PCNN models. The model is able to predict the correct relations with higher confidence as it able to gather evidence over multiple sentences in the bag. 

\subsection{Multi-instance Multi-label CNNs ~\citep{jiangrelation}}

The authors of this paper address the information loss problem in \citet{zeng2015distant} by using a cross-document max-pooling layer. 
Like in the attention model, they first find a vector representation, $r_i^j$ for each sentence $d_i^j$ of bag $i$. Then the final vector representation for the bag of sentences is found by taking a dimension wise max of the sentence vectors ($r_i^j, j=1,2...q_i$). The final feature vector for the bag of sentences is given as,

\begin{equation}
(r_i)_k = \max_{j=1,2...q_i} (r_i^j)_k; k=1,2...D
\end{equation}

where $r_i^j, r_i \in \mathbb{R}^{D}$. This allows each feature in the final feature vector to come from the most prominent document for that feature, rather than the entire feature vector coming from the overall most-prominent document.

The paper also address the issue of multi-label in relation extraction. Up until now, all models predicted a single relation class for an entity pair. But it is likely that the same entity pair can have multiply relations (called overlapping relations) which are supported by different documents. For example, (Steve\_Jobs, \texttt{Founded}, Apple) and (Steve\_Jobs, \texttt{CEO\_of}, Apple) are both valid relations between the same entity pair (Steve\_Jobs, Apple) which may be supported by different sentences. The authors modify the architecture to have sigmoid activation functions instead of softmax activations in the final layer, which would mean that the network predicts a probability for each relation class independently, rather than predicting a probability distribution over the relations. The loss for training the model is then defined as,

\begin{equation}
J(\theta) = \sum_{i=1}^T \sum_{r=1}^R y_r^i\text{log} p_r^i + (1-y_r^i)\text{log} (1-p_r^i)
\end{equation}

where $R$ is the number of relation classes, $p_r^i$ is probability for bag $i$ to have relation $r$ as predicted by the network and $y_r^i$ is a binary label if bag $i$ had relation $r$ or not.

The MIMLCNN model is able to improve performance of the PCNN and CNN models like the selective attention mechanism, as it is able to exploit the information across multiple documents in the bag, by using the most prominent document for each feature. The results are discuss further in the next section.

\section{Results}

Figure 3 summarizes the results of the various multi-instance learning models applied on the distant supervision dataset created by \citet{riedel2010modeling}. It shows the results for 3 non deep learning models namely \textit{Mintz} \citep{mintz2009distant}, \textit{MultiR} \citep{hoffmann2011knowledge} and MIML \citep{surdeanu2012multi}. We also see the performance of the deep learning models discussed in the previous sections.

It is observed that the all the deep learning models perform significantly better than the non deep learning models. Using the Multi-instance Multi-label (\textit{MIMLCNN}) mechanism with the CNN model further improves the curve over the \textit{PCNN} model. However, the selective attention mechanism applied over the \textit{PCNN} model gives the best performance out of all the models. It is interesting to note the increase in performance in the \textit{PCNN} curve to the \textit{PCNN+Att} curve as compared to the \textit{MIMLCNN} curve. Since the attention mechanism is a soft-selection mechanism, it works out to be more robust and able to exploit information across the sentences more effectively, than even the cross-document max mechanism used in \textit{MIMLCNN}. 

\begin{figure}[ht]
	\centering
	\includegraphics[width=0.45\textwidth]{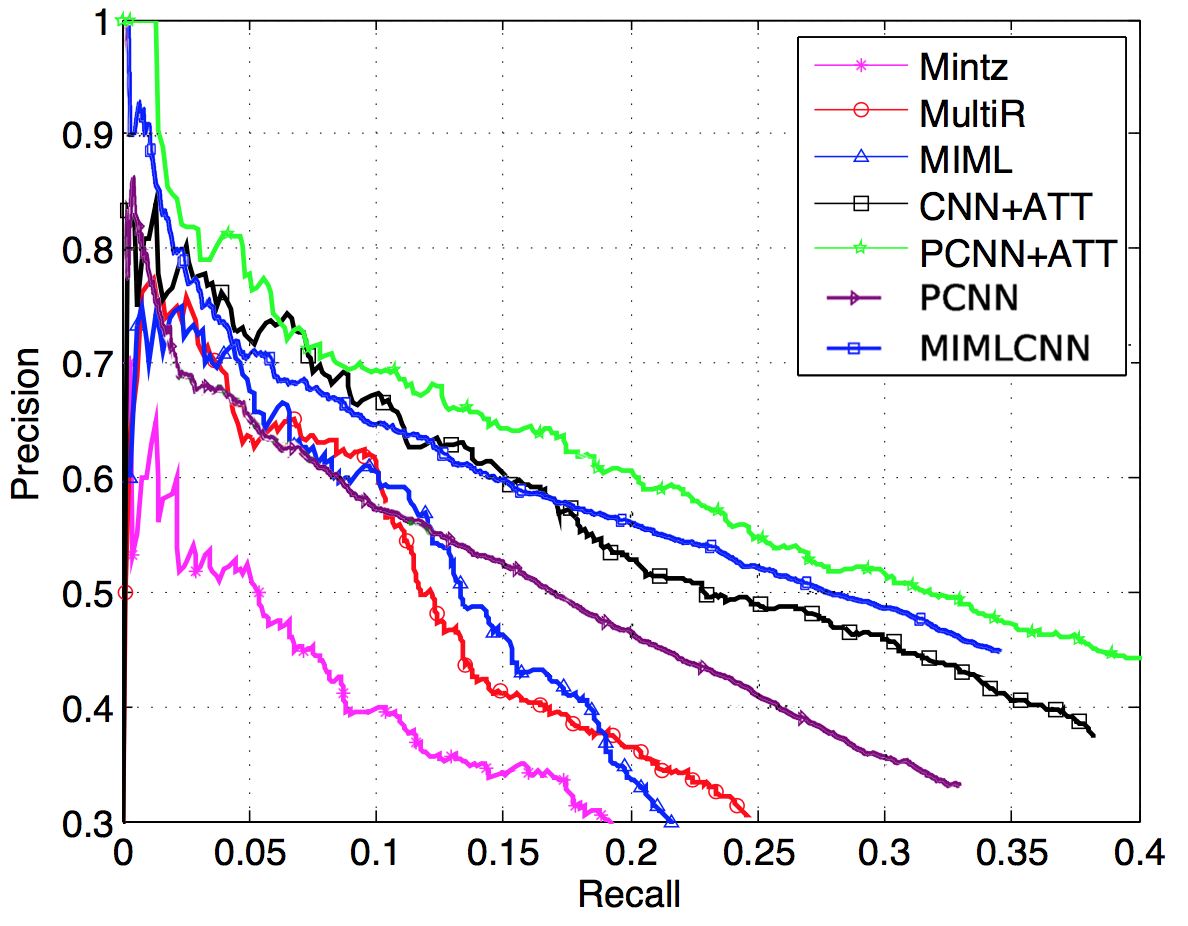}
    \caption{Results for Multi-instance learning models. (Sourced from \citep{lin2016neural} and \citep{jiangrelation})}
\end{figure}

\section{Concluding Remarks}

With the introduction of distant supervision for relation extraction by \citet{mintz2009distant}, modeling the task as Multi-instance problem has been widely adopted. Using this mechanism also provides enough data for deep learning models to be trained in the multi-instance setting which accommodates for the labeling noise in the data. Successive works have tried to handle the noise and distant supervision assumption with mechanisms like selective attention over document instances and cross-document max pooling, which have shown to increase performance. Some very recent works in the field also try to exploit the interaction between the relations by exploiting relation paths \citep{zeng2016incorporating} and relation class ties \citep{ye2016jointly} to improve the performance further. For example relations like \texttt{Father\_of} and \texttt{Mother\_of} can be exploited to extract instance for \texttt{Spouse\_of}. 

However these improvements only work on the training and inference methods of the model. As far as the deep learning aspect is concerned, the CNN or PCNN architecture used to encode the sentences is same across all these works. It is surprising to note that no work for the task of relation extraction has tried to use Recurrent Neural Networks (RNNs) for encoding the sentences in place of the CNNs (to the best of our knowledge). RNNs and LSTMs intuitively fit more naturally to natural language tasks. Even though NLP literature does not support a clear distinction between the domains where CNNs or RNNs perform better, recent works have shown that each provide complementary information for text classification tasks \citep{yin2017comparative}. Where RNNs perform well on document-level sentiment classification \citep{tang2015document}, some works have shown CNNs to outperform LSTMs on language modeling tasks \citep{dauphin2016language}. Future works for relation extraction can thus definitely try to experiment with using LSTMs for encoding sentence and relations.


\bibliography{acl2017}
\bibliographystyle{acl_natbib}

\end{document}